\documentclass{esannV2}
\usepackage{graphicx}
\usepackage{pgfplots}
\pgfplotsset{compat=newest}
\usepgfplotslibrary{groupplots}
\usepgfplotslibrary{dateplot}
\usepackage{inputenc}
\usepackage{amssymb,amsmath,array, multicol, subfig, wrapfig, multirow}
\expandafter\def\expandafter\normalsize\expandafter{%
	\normalsize
	\setlength\abovedisplayskip{3pt}
	\setlength\belowdisplayskip{3pt}
	\setlength\abovedisplayshortskip{3pt}
	\setlength\belowdisplayshortskip{3pt}
}

\usepackage{eso-pic}
\newcommand\AtPageUpperMyright[1]{\AtPageUpperLeft{%
		\put(\LenToUnit{0.5\paperwidth},\LenToUnit{-3cm}){%
			\parbox{1.45\textwidth}{\raggedleft\fontsize{9}{11}\selectfont #1}}%
}}%
\newcommand{\conf}[1]{%
	\AddToShipoutPictureBG*{%
		\AtPageUpperMyright{#1}
	}
}
\begin{document}
\title{Invariant Integration in Deep Convolutional Feature Space}

\author{Matthias Rath$^{1,2}$ and Alexandru P. Condurache$^{1,2}$
%
%
\vspace{.3cm}\\
%
1- University of L\"{u}beck - Institute for Signal Processing
%
\vspace{.1cm}\\
2- Robert Bosch GmbH - Automated Driving
}

\maketitle
\conf{\textsf{ESANN 2020 proceedings, European Symposium on Artificial Neural Networks, Computational Intelligence\\
		and Machine Learning. Bruges (Belgium), 2-4 October 2020, i6doc.com publ., ISBN 978-2-87587-074-2.}
}
\begin{abstract}
In this contribution, we show how to incorporate prior knowledge to a deep neural network architecture in a principled manner. We enforce feature space invariances using a novel layer based on invariant integration. This allows us to construct a complete feature space invariant to finite transformation groups.

We apply our proposed layer to explicitly insert invariance properties for vision-related classification tasks, 
demonstrate our approach for the case of rotation invariance and report state-of-the-art performance on the Rotated-MNIST dataset. Our method is especially beneficial when training with limited data.
\end{abstract}

\section{Introduction}
Deep neural networks (DNNs) are the state-of-the-art method to solve a wide variety of complex tasks from computer vision to speech recognition. In general, the weights of those networks are optimized exploiting a vast amount of data. However, collecting and especially labeling data, which is necessary for supervised learning, is time-consuming and labour-intensive.

Consequently, current research investigates the integration of prior knowledge to DNNs in order to improve the data efficiency of the learning process, thus improving the performance when training data is limited or scarce \cite{Coors2018ECCV}. 
One promising approach is to exploit prior knowledge by enforcing in- or equivariance properties to symmetric geometrical transformations such as rotation and scaling. This can be achieved by replacing the randomly initialized weights of convolutional layers with predefined or restricted filters \cite{HarmonicNetworks, DeepHybridNetworks, LearnableScatter}. Alternatively, the \textit{standard convolution}, which is equivariant to translations, can be substituted by a more general form called \textit{group convolution} \cite{GroupEquivariantCNNs}, which guarantees equivariance with respect to more sophisticated transformation groups. Weiler et al. combine both approaches by restricting the convolution weights to linear combinations of steerable filters and applying the group convolution to the activations \cite{LearningSteerable}. The aforementioned methods build invariant feature spaces by pooling from equivariant feature spaces. In practice however, those feature spaces are affected by sampling effects and are thus not completely invariant. To avoid this, one could replace the simple pooling operation with an advanced operation geared towards enforcing invariance properties.

We propose to use invariant integration (II) for this purpose. II is a method for constructing a complete feature space that is invariant to a transformation group \cite{SM_Existence,SM_Algos}. It has successfully been applied to object detection from grayscale images \cite{SM_Gray}, automatic speech recognition \cite{Muller3} and event detection from image sequences \cite{Condurache}. However, all of these approaches use II to create a feature space used by conventional classifiers, such as a \textit{Support Vector Machine}.

\begin{figure}
	\centering
	\includegraphics[width=0.8\textwidth]{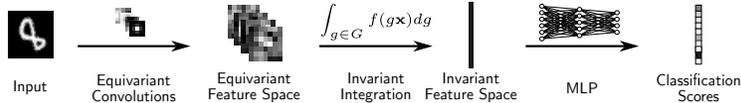}
	\caption{We apply invariant integration in a learned equivariant feature space to form a feature space invariant to rotations and translations and use fully connected layers for classification.}\label{fig:Architecture}
\end{figure}

We propose a novel DNN layer based on II embedded into a convolutional neural network (CNN) architecture. The II layer (IIL) builds upon an equivariant feature layer to foster an invariant feature representation (see Figure \ref{fig:Architecture}). We demonstrate that the backpropagation algorithm is applicable with respect to the IIL's parameters and input.
Hence, its parameters as well as preceding layers can be optimized en bloc.
We test our method on the Rotated-MNIST dataset \cite{RotMNIST} and demonstrate state-of-the-art performance. The largest performance gains are achieved when only few labeled training samples are available.

\section{A DNN architecture based on Invariant Integration}
Group theory lays the foundation of II. A group $G$ is a mathematical abstraction consisting of a set of elements acted upon by an operation under the axioms of closure, associativity, identity and invertibility. 
A function $f$ is said to be equivariant with respect to a group $G$ of transformations, if we can determine an exact relationship between transformations $g \in G$ of the function's input $x$ and a corresponding transformation $g^\prime \in G$ of the function's output:
\begin{equation*}
f(gx) = g^\prime f(x) \; \forall \; x \in X \text{.}
\end{equation*}
Invariance is a special case of equivariance, where the induced transformation in the output space $g^\prime$ is the identity. 

An example of an equivariant function is the convolutional layer, which is equivariant to translations. In practice, we are also interested to enforce invariance to other transformation groups than translations, e.g. rotations. However, this is more difficult to achieve in an explicit manner.

\setlength{\belowcaptionskip}{-15pt}

II is an algorithm to construct a complete feature space with respect to a symmetric transformation first proposed by Schulz-Mirbach \cite{SM_Existence, SM_Algos}. A feature space is defined as complete, if all patterns which are equivalent with respect to a transformation group $G$ are mapped to the same point in the feature domain, while all distinct patterns are mapped to different points. This means that a complete feature space is invariant to transformations $g\in G$ of the input signal.

In \cite{SM_Algos}, Schulz-Mirbach proposes a general algorithm to construct a complete feature space relying on the group average $A$, which is defined by integrating an invariant function $f$ over transformations $g \in G$ acting on the input space $\mathbf{x}$:
\begin{equation*}\label{eq:groupAverage}
A[f](\mathbf{x}) := \int_{g \in G} f(g\mathbf{x}) \,dg \text{.}
\end{equation*}
In the case of a discrete group, this integral reduces to a sum over all transformations $g \in G$ of the input.
For the invariant function $f$, the set of all possible monomials $m(\mathbf{x})$ represents a good choice for creating a complete feature space. These are defined as:
\begin{equation*}
m(\mathbf{x}) = \prod_{i=1}^{K} x_i^{b_i} \quad \text{with} \quad \sum_i b_i \le |G| \text{,}
\end{equation*}
where $K$ is the size of the input feature and $b_i$ are the monomial's exponents. The upper bound for the number of all possible monomials is $\binom{K + |G|}{K}$.

\subsection{Rotation-Invariant Integration in Image Space}
We apply II using monomials over the group of rotations and translations in image space 
\cite{SM_Gray}. Thus, our input is defined as $\mathbf{x} \in \mathbb{R}^{H \times W \times C}$, where $H$, $W$ and $C$ are the image's width, height and number of channels. We apply the group average independently to each channel $c = 1,\hdots,C$ and reformulate:
\begin{align*}
A[f](\mathbf{x}) &= \sum\limits_{u}\sum\limits_{v}\int_\phi m(\mathbf{x}(u, v; \phi))\,d\phi \\
&= {\sum\limits_{u}\sum\limits_{v}}\int_\phi\prod\limits_{i=1}^{K}\mathbf{x}(u + d_{u,i}\sin(\phi), v + d_{v,i}\cos(\phi))^{b_i}\,d\phi \text{,}
\end{align*}
with the monomial distances $d_{u,i}$ and $d_{v,i}$, the pixel offsets $u$ and $v$ and the integration angle $\phi$.
The output is of dimension $\mathbb{R}^{C \times M}$, where $M$ is the number of monomials. Contrarily to previous work, we choose to apply rotation-invariant integration directly in convolutional feature space.

\subsection{Monomial Selection}
Since including all possible monomials is computationally expensive, the question arises, how to choose a meaningful subset. While the exponents $b_i$ can be optimized using backpropagation (see Chapter \ref{sec:backprob}), choosing the monomial orders $K$ and their distances is non-trivial.

We use an iterative approach for feature selection called \textit{Feature Finding Neural Network} (FFNN) proposed by Gramss \cite{Gramss91}. It has successfully been used to select the monomials for II in automatic speech recognition by Mueller and Mertins \cite{Muller3} and is based on iteratively ranking monomial combinations according to the \textit{least square error} (LSE) of a linear classifier. While using a linear classifier facilitates good generalization ability, the LSE can be computed in closed-form, which enables an efficient calculation.

We adapt this method by utilizing the closed-form solution of a linear classifier to rank the feature combinations according to the classifiers' validation accuracies instead of the LSE.
The iterative selection is stopped, if the validation accuracy has not improved within the past ten iterations. 

\subsection{Architecture and Training Process}
The architecture for a classification network using the proposed IIL is shown in Figure \ref{fig:Architecture}.
The IIL is used to construct an invariant feature space with respect to transformations of the convolutional feature space it acts on. To ensure that the whole DNN is invariant, the convolutional feature space must be equivariant with respect to transformations of the DNN's input. Thus, we use an equivariant backbone built out of equivariant convolutional layers and apply the IIL to build an complete invariant feature space (see Figure \ref{fig:Architecture}). Finally, we use fully connected layers to obtain the classification scores. In comparison, the networks proposed in \cite{HarmonicNetworks, LearningSteerable} use max- or sum-pooling to create an invariant feature space from their equivariant convolutional feature space.

During training, we first train the baseline network without the IIL. Thereafter, we apply II on the features of the last convolutional layer and select $M$ monomials using our iterative sampling approach. Finally, we re-train the entire network including the IIL and the preceding equivariant convolutional layers. 

\subsection{Backpropagation for Invariant Integration Layers}\label{sec:backprob}
To enable using II for the construction of a complete feature space in combination with DNNs it is inevitable that the proposed feature transformation is differentiable with respect to both its parameters and inputs.
The gradients of a monomial with respect to a single input $x_j$ and to the learnable exponents $b_j$ are respectively defined as:
\begin{equation*}
\frac{\delta m(\mathbf{x})}{\delta x_j} = b_j x_j^{b_j - 1} \prod_{i=1, i\neq j}^{K}x_i^{b_i}
\text{,} \quad
\frac{\delta m(\mathbf{x})}{\delta b_j} = \log(x_j) x_j ^{b_j} \prod_{i=1, i\neq j}^{K}x_i^{b_i} \text{.} 
\end{equation*}
Investigating these formulas, it is evident that we need to enforce $x_i \neq 0 \; \forall i$,
because 
$\lim\limits_{x_i \rightarrow 0} \frac{\delta m(\mathbf{x})}{\delta x_j} = 0$ and $\lim\limits_{x_i \rightarrow 0} \frac{\delta m(\mathbf{x})}{\delta b_j} = 0$,
would lead to vanishing gradients backpropagated to earlier layers and to the exponents of the IIL, respectively. 

We further restrict the input values to $\mathbb{R}^{+}$ because multiplying by a negative number alters the sign of the output, hence leading to great shifts in output space. Finally, we note that the identity element of a multiplication is 1, while a multiplication with values close to 0 would eradicate the result. Following the aforementioned insights, we choose to shift our input features:
${\mathbf{\tilde{x}} = \max(\epsilon, \mathbf{x} - x_{min} + 1)} \text{,}$
with $0 < \epsilon \ll 1$. In our implementation, we shift the input per channel, while $x_{min}$ is the minimal value that occurred within that channel when processing the training data. Lastly, we apply the $\max$-function to ensure that there are no values smaller than $\epsilon$ during testing.

\section{Experiments and Discussion}
\setlength{\belowcaptionskip}{-15pt}
\begin{figure}
	\begin{multicols}{2}
		\includegraphics[width=\linewidth]{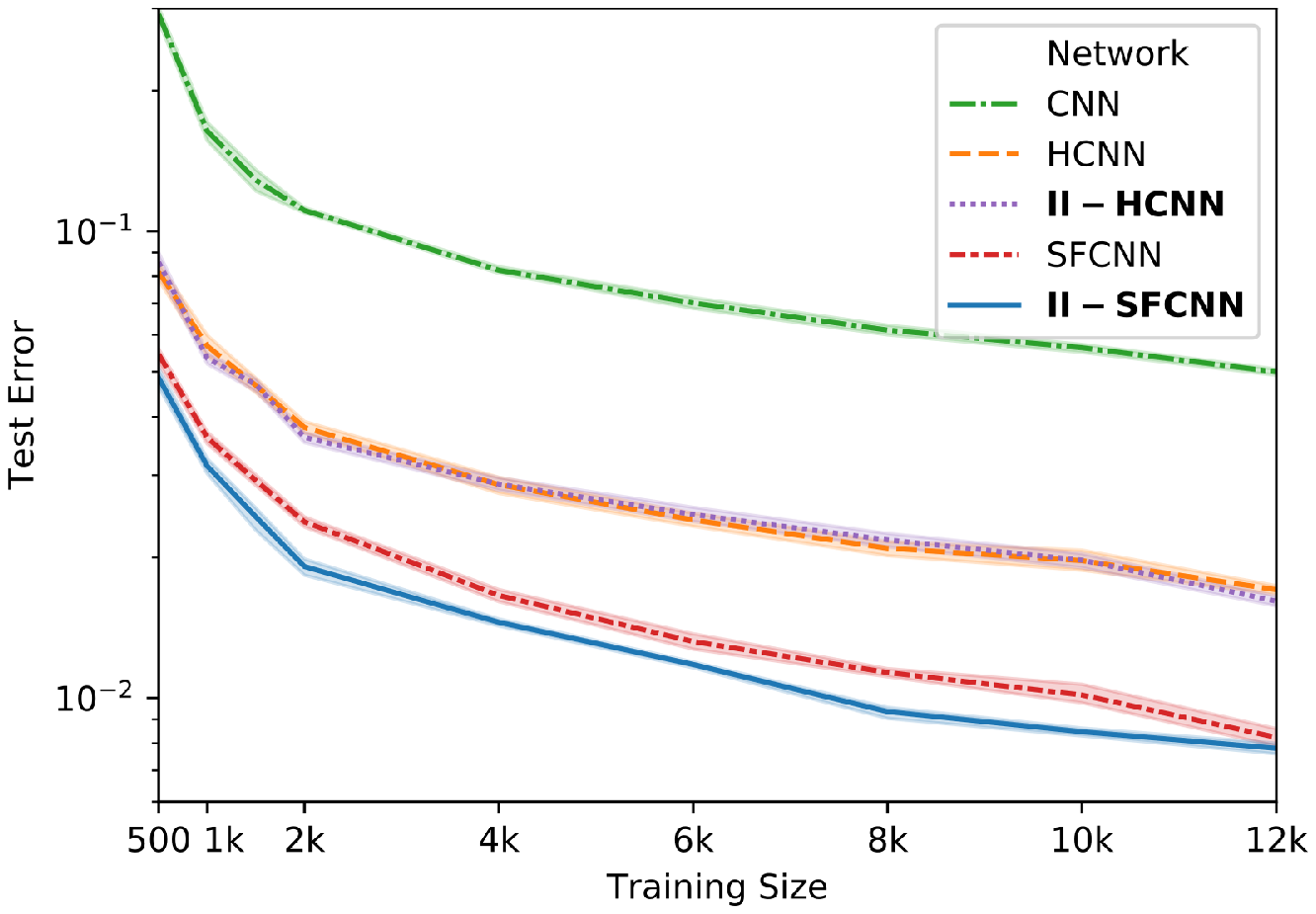} \par
		\includegraphics[width=\linewidth]{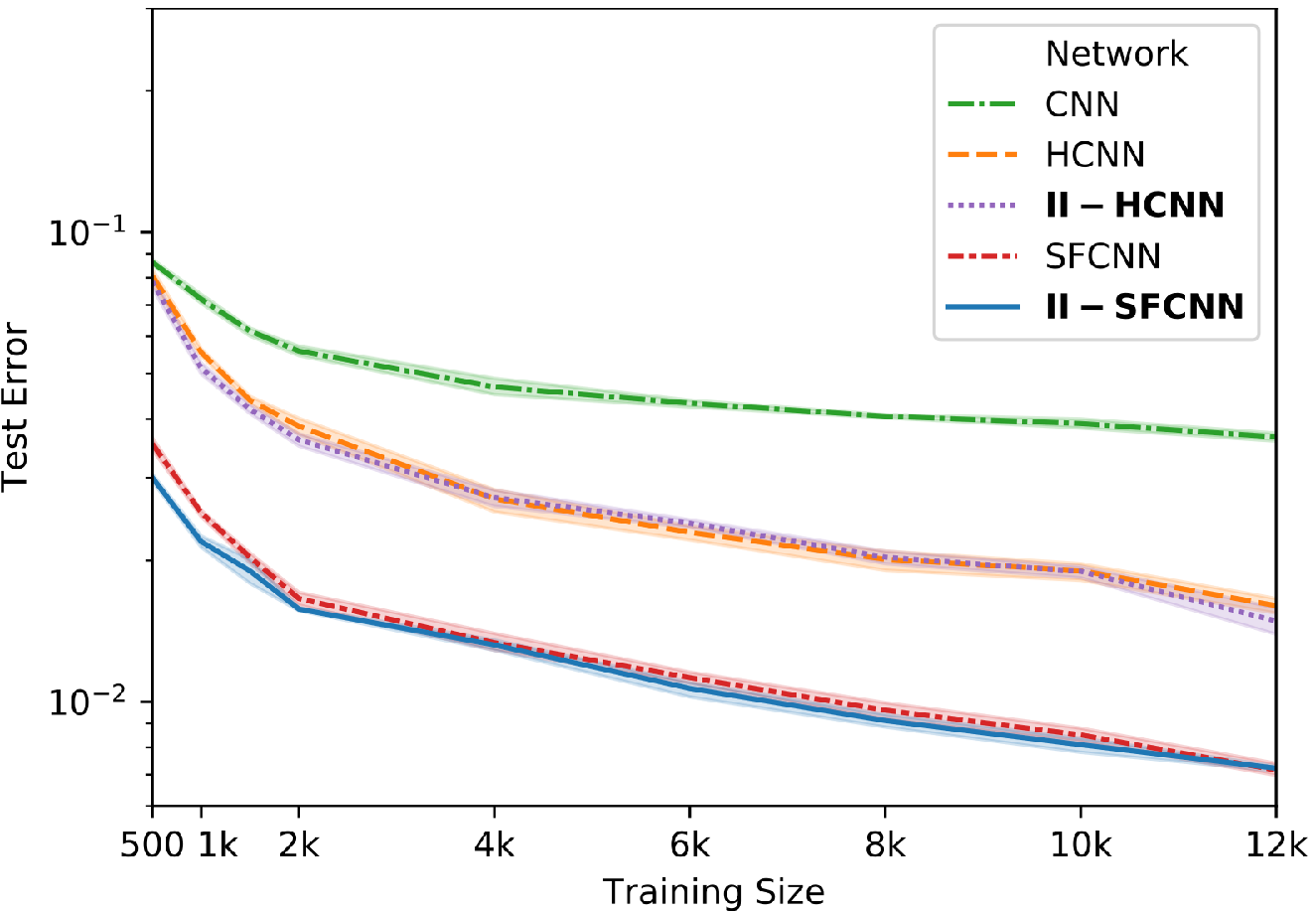}\par 
	\end{multicols}
	\centering
	\vspace*{-10pt}
	\caption{Mean test error using limited training data without (\textit{left}) and with (\textit{right}) augmented data. The transparent corridors represent the results' standard deviation.}
	\label{fig:MNIST_limited}
\end{figure}

We use our approach to enforce rotation invariance for the recognition of randomly rotated hand-written digits on the Rotated MNIST dataset \cite{RotMNIST}. It contains $10{,}000$ training, $2{,}000$ validation and $50{,}000$ test images. We optimize the hyper-parameters on the validation set.
Thereafter, we retrain on the training and validation sets to report the final results on the test set (cf. \cite{HarmonicNetworks, LearningSteerable}).

\setlength\intextsep{0pt}
\setlength{\belowcaptionskip}{0pt}
\begin{wraptable}{R}{4.8cm}
	\small
	\centering
	\begin{tabular}{|l|c|c|} 
		\hline
		\multirow{2}{*}{Method} & \multicolumn{2}{c|}{Augmentation} \\ 
		& $\times$ & $\checkmark$ \\
		\hline
		CNN \cite{HarmonicNetworks,GroupEquivariantCNNs} & $5.130$ & $3.772$\\ 
		HCNN \cite{HarmonicNetworks} & $1.730$ & $1.606$ \\ 
		\textbf{II-HCNN}& $1.548$ & $1.377$ \\ 
		SFCNN \cite{LearningSteerable} & $0.880$ & $\mathbf{0.714}$ \\ 
		\textbf{II-SFCNN} & $\mathbf{0.799}$ & $0.722$ \\ 
		\hline
	\end{tabular}
	\caption{Mean test error [$\%$].}
	\label{tab:results}
\end{wraptable}

For the equivariant backbone, we use two different approaches that are equivariant to rotations in image space. First, we apply harmonic convolutions (HCNN) proposed by Worrall et. al \cite{HarmonicNetworks}. We use the modulus of the final complex convolutional layer as the input to our IIL. Secondly, we use the Steerable Filter CNN (SFCNN) proposed by Weiler et. al \cite{LearningSteerable}. We max-pool along the angle dimension of the final convolutional layer and apply channel-wise II. 
We empirically choose to use $M=5$ monomials and to integrate over 8 angles using bilinear interpolation to obtain a good trade-off between performance and computational complexity.

We compare the different approaches by means of the mean test error over ten independent training runs. In Table \ref{tab:results}, we show results obtained when training on the entire training and validation set both with and without data augmentation. We evaluate a standard CNN as well as the HCNN and the SFCNN without and with the IIL. The data was augmented by randomly rotating the input during training.
When combining II with the HCNN, we achieve a relative improvement of $10.52 \, \%$ without and $14.26\, \%$ with data augmentation. The II-SFCNN improves the baseline by $9.2 \, \%$. However, when training with augmented data, the SFCNN achieves the best performance, but only by a small margin.

Figure \ref{fig:MNIST_limited} shows the mean test error and the standard deviation when training with limited data. When using only a subset of the available training data, we achieve significant improvements over the state-of-the-art baseline algorithms. It seems that enforcing the sought invariances on top of an equivariant feature space by using II instead of the simpler pooling operation is particularly beneficial when only limited data is available. We believe this is related to the fact that a small training set typically does not include the variability needed to properly capture the sought invariances without explicitly enforcing them. 
Additionally, we observe that the performance boost is bigger when applying the IIL to the SFCNN because its feature representation is equivariant to sampled rotations, while the HCNN guarantees equivariance to continuous rotations. Enforcing the invariance with the IIL seems to mitigate those sampling effects.

\section{Conclusion and Future Work}
We proposed a novel layer based on II which allows us to incorporate prior knowledge expressed as the need to obtain features invariant to certain geometrical transformations of the input, e.g. rotations. We apply it to enforce rotation invariance in a CNN architecture. 
We show competitive performance when training on the full dataset and state-of-the-art results in the low data regime. This indicates that our approach enables a better transition from equivariant to invariant features than pooling when the learned feature representation is not exactly equivariant, e.g. due to sampling effects. Our method is especially helpful, when training with fewer training samples.

In the future, II can be expanded towards other transformation groups, e.g. scale. Additionally, the IIL can be applied in more complex neural network architectures. We will investigate, if hand-designing the monomials is possible, thus allowing to incorporate additional prior knowledge and avoiding the pre-training of the baseline network necessary to enable selecting appropriate monomials.



\begin{footnotesize}

\bibliographystyle{unsrt}
\bibliography{egbib}

\begin{thebibliography}{10}

\bibitem{Coors2018ECCV}
B.~Coors, A.~P. Condurache, and A.~Geiger.
\newblock Spherenet: Learning spherical representations for detection and
  classification in omnidirectional images.
\newblock In {\em ECCV 2018}, pages 525--541.

\bibitem{HarmonicNetworks}
D.~E. Worrall, S.~J. Garbin, D.~Turmukhambetov, and G.~J. Brostow.
\newblock Harmonic networks: Deep translation and rotation equivariance.
\newblock In {\em CVPR 2017}, pages 7168--7177.

\bibitem{DeepHybridNetworks}
E.~Oyallon, E.~Belilovsky, and S.~Zagoruyko.
\newblock Scaling the scattering transform: Deep hybrid networks.
\newblock In {\em ICCV 2017}, pages 5619--5628.

\bibitem{LearnableScatter}
F.~Cotter and N.~G. Kingsbury.
\newblock A learnable scatternet: Locally invariant convolutional layers.
\newblock In {\em {ICIP} 2019}, pages 350--354.

\bibitem{GroupEquivariantCNNs}
T.~Cohen and M.~Welling.
\newblock Group equivariant convolutional networks.
\newblock In {\em ICML 2016}, pages 2990--2999.

\bibitem{LearningSteerable}
M.~Weiler, F.A. Hamprecht, and M.~Storath.
\newblock Learning steerable filters for rotation equivariant cnns.
\newblock In {\em CVPR 2018}, pages 849--858.

\bibitem{SM_Existence}
H.~Schulz-Mirbach.
\newblock On the existence of complete invariant feature spaces in pattern
  recognition.
\newblock In {\em ICPR 1992}, pages 178 -- 182.

\bibitem{SM_Algos}
H.~Schulz-Mirbach.
\newblock Algorithms for the construction of invariant features.
\newblock In {\em DAGM-Symposium 1994}, pages 324--332.

\bibitem{SM_Gray}
H.~Schulz{-}Mirbach.
\newblock Invariant features for gray scale images.
\newblock In {\em DAGM-Symposium 1995}, pages 1--14.

\bibitem{Muller3}
F.~M{\"{u}}ller and A.~Mertins.
\newblock Contextual invariant-integration features for improved
  speaker-independent speech recognition.
\newblock {\em Speech Communication}, 53(6):830--841, 2011.

\bibitem{Condurache}
A.~P. Condurache and A.~Mertins.
\newblock Sparse representations and invariant sequence-feature extraction for
  event detection.
\newblock {\em VISAPP 2012}, pages 679--684.

\bibitem{RotMNIST}
H.~Larochelle, D.~Erhan, A.~C. Courville, J.~Bergstra, and Y.~Bengio.
\newblock An empirical evaluation of deep architectures on problems with many
  factors of variation.
\newblock In {\em ICML 2007}, pages 473--480.

\bibitem{Gramss91}
T.~{Gramss}.
\newblock Word recognition with the feature finding neural network (ffnn).
\newblock In {\em NNSP 1991}, pages 289--298.

\end{thebibliography}

\end{footnotesize}
\end{document}